\pdfoutput=1  
\documentclass[letterpaper, 10 pt, conference]{ieeeconf}  

\IEEEoverridecommandlockouts                              

\usepackage{cite}
\usepackage{amsmath,amssymb,amsfonts,bm}
\usepackage{graphicx}
\usepackage{textcomp}
\usepackage{xcolor}
\usepackage{booktabs}
\usepackage{multirow}
\usepackage{array}
\usepackage{float}
\usepackage{algorithm}
\usepackage{algpseudocode}
\usepackage{siunitx}
\usepackage{url}
\usepackage[utf8]{inputenc}

\title{\LARGE \bf
A Reproducible Reconstruction and Evaluation Framework for Photorealistic Lunar Analog Scene Reconstruction Toward Digital Twins
}

\author{Antar Mazumder$^{1*}$, Ryan Hartzell$^{1}$, Luke Boyd$^{1}$, Alex Robinson$^{1}$\\
Neha Sharma$^{2}$, Chance Carnival$^{1}$, and Frances Zhu$^{1}$%
\thanks{\copyright{} 2026 IEEE. Personal use of this material is permitted.
Permission from IEEE must be obtained for all other uses, in any current or
future media, including reprinting/republishing this material for advertising
or promotional purposes, creating new collective works, for resale or
redistribution to servers or lists, or reuse of any copyrighted component of
this work in other works.}%
\thanks{$^{1}$Dept. of Mechanical Engineering, Colorado School of Mines, Golden, CO, USA}%
\thanks{$^{2}$Dept. of Eletrical Engineering, Colorado School of Mines, Golden, CO, USA}%
\thanks{$^{*}$Corresponding author: Antar Mazumder {\tt\small antar\_mazumder@mines.edu}}%
}

\begin{document}

\maketitle
\thispagestyle{empty}
\pagestyle{empty}

\begin{abstract}
 In this paper, we present a reconstruction pipeline that empirically quantifies the photorealistic scene reconstruction accuracy from both human-controlled and rover mast camera monocular feeds against physical ground truth as a function of Lunar traversal effects (e.g., motion blur, haze due to lofted regolith, trajectory bias, and sparse views). First, we benchmark six variants of 3D Gaussian Splatting (3DGS) across three Lunar analog scenes on held-out test sets using standard photometric metrics. We then establish how under traversal effects, this test-set-specific Novel View Synthesis (NVS) performance does not proportionally scale to the same photometric metrics in the reconstructed DTs. Our results quantify that splat training metrics alone are insufficient predictors of Cross-Reconstruction Photorealistic NVS Fidelity (CRPNVS), and that factors such as rover trajectory bias and geometric scene coverage play a critical role in determining the photorealistic quality of the reconstructed scenes targeting Lunar analog Digital Twins. Additionally, we pilot study Coverage Score ($S_{c}$), a geometry-based diagnostic derived from COLMAP\cite{colmap} poses and point cloud structure that provides a meaningful heuristic about the highest achievable CRPNVS structural fidelity, independently of splat training performance in the absence of a human-controlled reference. These findings establish a standardized, human control-referenced evaluation pipeline for photorealistic scene reconstruction for Lunar-analog Digital Twin development and future planetary exploration.
 
Code and project page:
\url{https://rose-lab-admin.github.io/rose-lunar-dt-rec}.

\end{abstract}

\section{Introduction}

Collecting robot point-of-view (PoV) data from the target or extremely similar environments is crucial for training and validating vision-based algorithms or policies. However, doing so for extraterrestrial missions poses significant challenges as the target venues such as Lunar terrains are simply not as accessible as their terrestrial counterparts. Contemporary methods prevalent in large volumes of literature indicate three approaches for addressing this obstacle via: (1) natural analog environments \cite{vayugundla2018datasets,arm2023scientific}, (2) artificial surface simulator testbed facilities \cite{knapmeyer2025first} and (3) high fidelity virtual testbeds, such as Digital Twins (DTs)\cite{richard2024omnilrs}. Disregarding the sim2real gap momentarily, DTs have recently emerged as a potential alternative in circumventing the challenges of physical analog environments and testbed facilities (e.g., the replication and manipulation of environmental variables such as gravity, lighting, wind speed, and granular media, and the requirement of a physical robot). Such advantages have made DTs a convenient source of synthetic training data for image-based algorithms. 

However, considering the sim2real gap, purely synthetic data might relate very poorly to real-world scenes\cite{hao2024synthetic,karoly2023increasing}. Therefore, photorealistic Lunar-analog scene reconstructions targeting Digital Twins need to have a strong basis from the target physical environment (the input fidelity criterion), which dictates photorealism of the synthetic data derived from these reconstructions (the output fidelity criterion). Furthermore, these criteria need to be not only qualified but also quantified by relevant metrics such as Structural Similarity Index Measure (SSIM), Peak Signal-to-Noise Ratio (PSNR), and Learned Perceptual Image Patch Similarity (LPIPS) to provide insight into the reliability of the DT in question for developing robust vision algorithms. However, this reconstruction and validation practice that quantifies ground truth vs photorealistic scene reconstruction toward Lunar-analog Digital Twins using direct photometric metrics under Lunar traversal effects remains largely absent in available research, as will be discussed in \textit{Related Works}.

\begin{figure}
    \centering
    \includegraphics[width=1\linewidth]{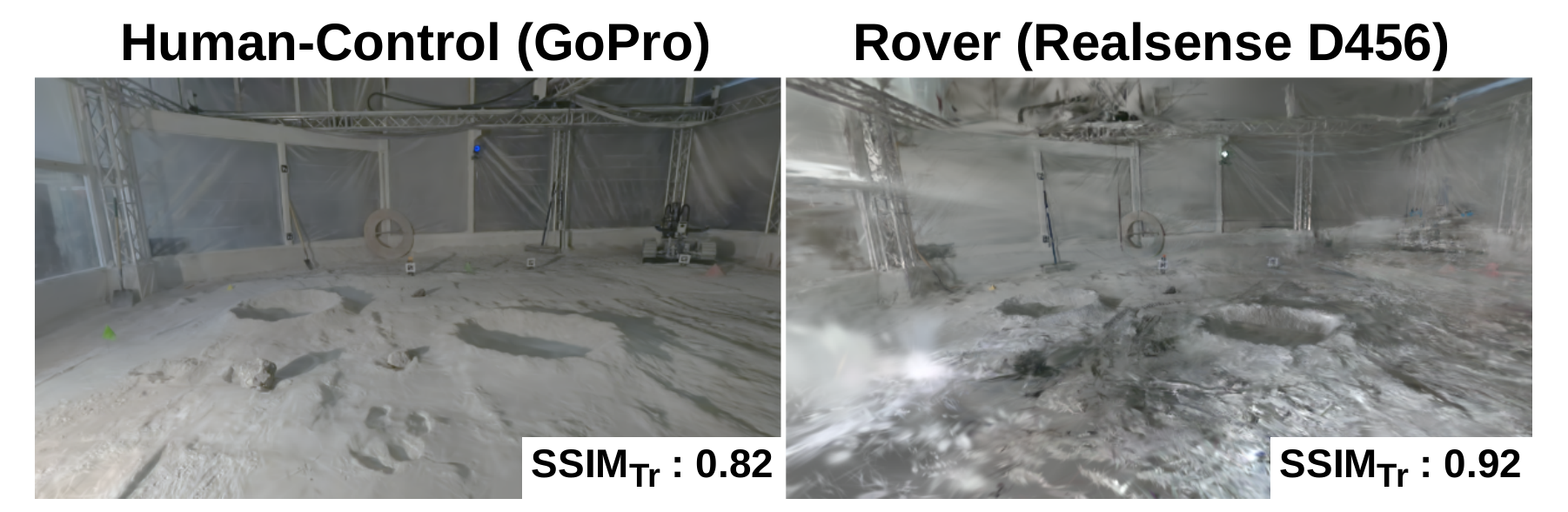}
    \caption{Under traversal effects, high training metrics on held-out test images fail to guarantee similar level of novel view synthesis performance across the entire reconstructed testbed.}
    \label{fig1:nvs_issue}
    \vspace{-1.5em}
\end{figure}

\subsection{Related Works}
A significant volume of literature is aimed at collecting Lunar-analog data without pursuing photorealistic reconstruction. For example, in\cite{vayugundla2018datasets}, the authors only collect and present visual and sensor data on Mount Etna (Natural Lunar analog). From artificial testbeds, the POLAR dataset \cite{wong2017polar} provides stereo imagery from NASA Ames, SPICE-HL3 \cite{rodriguez2026spice} covers stereo, monochrome, and single‑photon avalanche diode modalities from LunaLab at the University of Luxembourg. The authors in \cite{martens2024synthetic} offer RGB and event-camera streams from Exterres Lab. None of these expands toward photorealistic reconstruction or reports a photometric assessment of their visual data.

 Another prominent group addresses scene reconstruction but stops short of direct photometric evaluation. For instance, the authors in \cite{hansen2024polar} extend the POLAR dataset to COLMAP-based\cite{colmap} dense reconstruction under simulated Lunar light but report only pixel-level reconstruction error, not photometric metrics. POLAR-Sim \cite{chen2025polar} generates high-fidelity meshes with photometric texturing, yet only qualitatively claims visual consistency. OmniLRS \cite{richard2024omnilrs} evaluates photorealism through a downstream rock segmentation performance, which is not a direct photometric measurement. Additionally, some textures are sourced outside the target environment, potentially reducing true environmental accuracy.
 
 There has been a very recent surge in photorealistic Lunar scene reconstruction using 3D Gaussian Splatting (3DGS) \cite{kerbl20233d} and its variations, most of which do report photometric metrics. For example, the authors in \cite{zhang20253d} benchmark Gaussian Splat variants on SPICE-HL3 in terms of SSIM, PSNR and LPIPS on held-out test sets.
 In \cite{prosvetov2025illuminating}, the authors compare photogrammetry, NeRF, and Gaussian Splatting across SPICE-HL3, NASA LOLA, and synthetic data in terms of similar photometric metrics. The study in\cite{casadesus2026semantic} pursues large-scale Lunar mapping via real-time 3DGS fused with stereo depth on synthetic LuPNT data. However, these approaches have three distinct flaws: (1) leveraging held-out test-set metrics that might inflate with trajectory bias, resulting in poor generalization across the entire reconstruction; (2) not discussing change in method rankings under Lunar-traversal effects; and (3) not quantifying coverage sparsity affecting photorealistic performance.



\begin{figure*}
    \centering
    \includegraphics[width=0.93\linewidth]{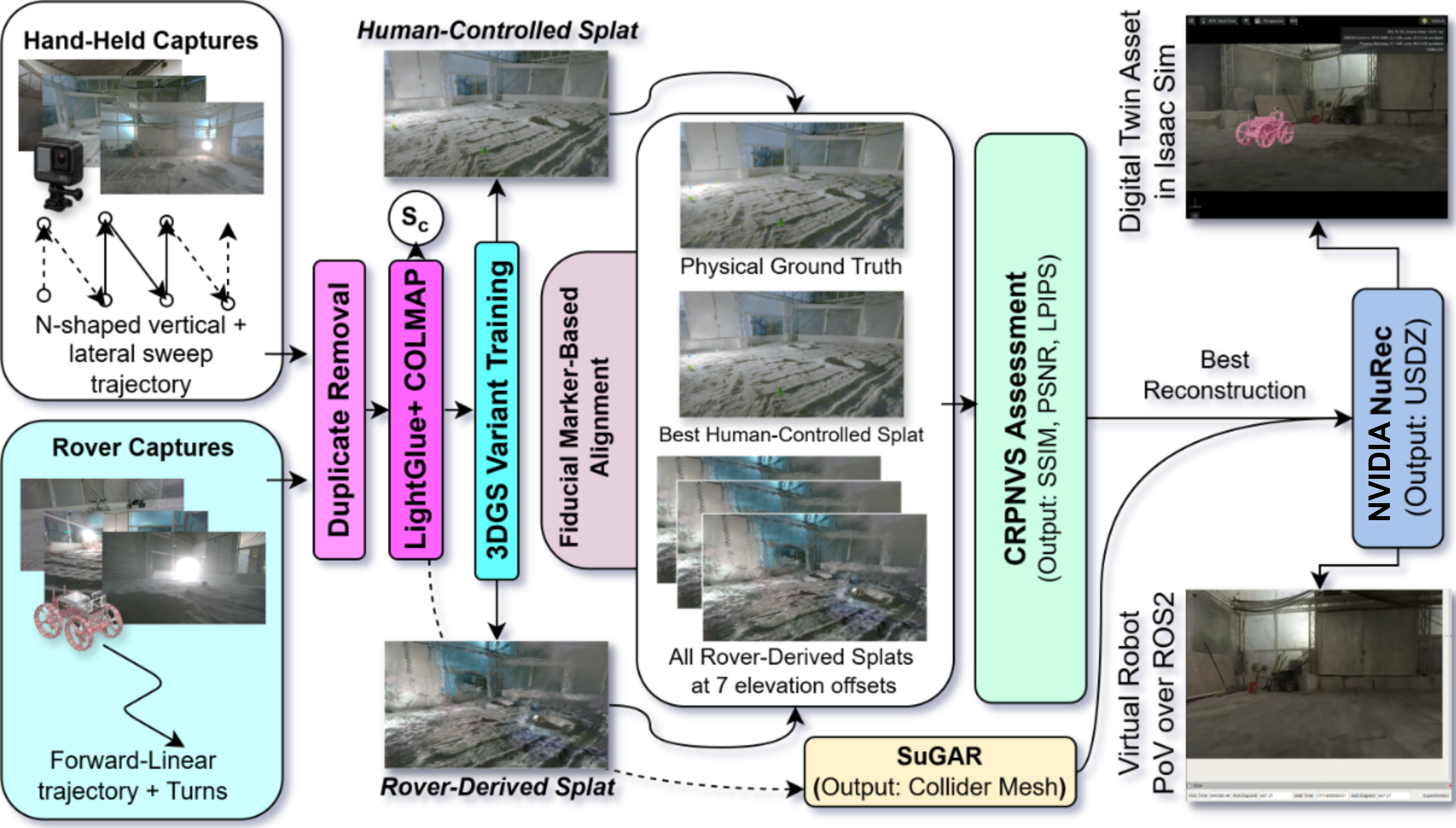}
    \caption{End-to-end photorealistic scene reconstruction and evaluation pipeline from data collection to simulation-ready Lunar-analog DT asset generation for Isaac Sim 5. \textbf{CRPNVS:} Cross-Reconstruction Photorealistic Novel View Synthesis; $\mathbf{S_c}$ denotes Coverage Score.}
    \label{fig:placeholder}
    \vspace{-1.5em}
\end{figure*}



\subsection{Research Gaps}
From the explored literature, five gaps are identified:

\begin{enumerate}
    \item \textit{\textbf{Absent or indirect photorealism evaluation:}} Most works
    either omit photorealistic Lunar-scene reconstruction for DTs entirely or claim
    photorealism without direct photometric measurement against a physical
    ground truth (GT).

    \item \textit{\textbf{Training metrics conflated with reconstruction fidelity:}} Surveyed
    studies using photometric metrics depend solely on held-out test images during training. However, under traversal effects these metrics
    do not relate to Novel-View-Synthesis (NVS) quality at all viewpoints (Fig.\ref{fig1:nvs_issue}). For example, if the camera trajectory is mostly linear (i.e. trajectory bias), most images are zoomed in/out versions of each other, which causes model overfitting. In such cases, training metrics alone are
    unreliable indicators of true photorealistic reconstruction fidelity.

    \item \textit{\textbf{No human-controlled vs. rover fidelity comparison:}} Although standard practice of terrestrial DT reconstruction involves human-controlled data, in frontier space exploration where this control is limited, rover data (autonomous or teleoperated) might be the only source to quantify for photoreal scene reconstruction. No work in
    the surveyed literature quantifies the photorealistic reconstruction gap
    between human-controlled input and rover-derived imagery which undergoes traversal effects against a physical ground
    truth.

    \item \textit{\textbf{Unquantified sparsity:}} Studies acknowledging coverage sparsity as a
    degradation factor in Lunar scene reconstruction do not quantify coverage
    levels affecting the metrics, making it difficult to pre-diagnose reconstruction fidelity from a
    given dataset.

    \item \textit{\textbf{No reproducible standard pipeline:}} No reproducible reconstruction and
    evaluation framework exists in the surveyed literature that discusses reconstruction fidelity as a function of the nuances
    identified above.
\end{enumerate}
\subsection*{Contributions}
The primary contribution of this paper is a reproducible, highly photorealistic scene reconstruction and validation pipeline that guides the readers from data collection to obtaining simulation-ready Isaac Sim 5 assets for Lunar-analog DTs using only handheld monocular image captures. In addition, it replicates the same pipeline from rover-derived imagery and uses a cross-reconstruction photorealistic novel view synthesis (CRPNVS) assessment method to: (1) establish why under Lunar-traversal effects (trajectory bias, sparsity, dusty PoVs, and motion blur), training metrics from held-out test-sets are flawed fidelity indicators, (2) properly quantify scene-specific fidelity loss under the traversal effects, and (3) highlight expected and realistic rover-based fidelity loss against the proposed human-controlled reconstruction and ground truth. 
To compensate for datasets that have traversal effects but not a controlled or ground truth baseline, this work proposes, evaluates and justifies a Coverage Score, $S_c \in [0,1]$, derived purely from COLMAP~\cite{colmap} poses and point cloud geometry as a pre-diagnostic tool for expected photorealism from a given set of images. 



\section{Method}
This section discusses the experimental settings, data collection pipelines, Gaussian Splat variants explored, the proposed coverage score, alignment methods and how they all come together to form a cross-reconstruction evaluation framework. 
\subsection{Testbed and Sensors}

\paragraph{Testbed}
Experiments are conducted on the Mines Lunar Surface Simulator (MLSS) at 
the Colorado School of Mines. It is a \SI{108}{\metre\squared} Lunar-regolith simulant testbed designed to simulate Lunar highland regions using CSM-LHT-T type 
simulant~\cite{jehn2025implications}. The testbed supports configurable terrain features 
including mounds, craters, and rocks with a set of ArUco markers 
fixed to its walls. Within the MLSS, each experiment is conducted over a 
defined test zone, the area of which vary between
experimental configurations. For highly dusty scenes (Rover-Scene 3), four additional 
AprilTags are placed closer to the test zone. These 
fiducial markers are used for splat alignment across modalities, rover 
localization, and scale error estimation.

\paragraph{Sensors \& Data Collection}
Two sensors are used for collecting image data: human-control uses a stabilized GoPro Hero~8 
camera that captures images at 4K resolution and 60~FPS. It follows a ``\textbf{N-shaped}'' sweep motion around the test zone 
to achieve optimal vertical and horizontal coverage, with significant 
emphasis on ground-level viewpoints. The camera experiences minimal to no rotation around its axes and always faces the front direction. For easier intuition, readers are encouraged to consider an imaginary wall surrounding the test-zone, where the camera trajectory paints a chain of the letter ``\textbf{N}'' on the outer surfaces of the wall. The bottom edge of ``\textbf{N}'' touches the ground plane as close as possible. This trajectory ensures good scene overlap while minimizing duplicate frames by avoiding forward-linear motion. The second sensor is an Intel 
RealSense D456 RGB-D camera, which serves as a fixed front-facing Mast Camera 
mounted on a rover, RoSey, based on Astrobotic's 6U CubeRover 
specification~\cite{astrobotic_cuberover}. The derived feed is $848 \times 480$ pixels with an 8-bit color depth obtained at roughly 12 FPS. It is not confined to any fixed trajectory, rather teleoperated to get close inspections of regions of interest, such as craters and rocks.
\subsection{Scenes}
Three pairs of different Lunar scenarios are simulated. Scene~1 contains a 
diffused light environment relevant to indoor Lunar bases or construction 
sites. Scene~2 features low-horizon Lunar south-pole-like 
lighting with rocks and craters. Scene~3 shares the same
lighting but introduces variation in crater locations. For the human-control, 
each scene is scanned as-is to provide  clean physical ground truth PoVs. The
reconstructions from the human-controlled scans are denoted as
$\mathcal{R}_{\mathrm{Ctrl},i}$, where $i = 1, 2, 3$. The mast camera feed, however, undergoes
rover-motion-induced degradation including motion blur and lofted regolith
dust. The rover traversing Scene~3 is particularly challenged, with extreme and varying
regolith dust levels to mimic a very difficult reconstruction task. Rover-derived reconstructions are denoted as $\mathcal{R}_{\mathrm{Rov},i}$, where $i = 1, 2, 3$. These conditions
ensure that the rover-based reconstructions capture realistic traversal effects
compared to the undisturbed $\mathcal{R}_{\mathrm{Ctrl},i}$.

\subsection{Data Processing}
The total number of sampled images is roughly 70,000 frames 
for the Control (GoPro) and around 5,000 frames for the mast camera. To reduce 
this load, first a FPS-based temporal sampling is applied to down-select frames 
at a uniform rate, followed by perceptual hash-based duplicate 
removal~\cite{zauner2010implementation} which eliminates roughly 
15--20\% of GoPro images and more than 70--80\% of mast camera images. This higher percentage of duplicate removal for the rover data indicates significant trajectory bias. 
Next, based on sharpness criteria, the 1,000 best 
images are selected. These images undergo a LightGlue-based 
COLMAP\cite{colmap}~\cite{lindenberger2023lightglue} reconstruction, which registers a subset of input images based on detected features. This work uses Depth Anything 3 
(DA3)~\cite{lin2025depth}, a state-of-the-art Monocular Depth 
Estimation (MDE) method, to generate corresponding depth maps for the input 
images. It relies on MDE for two reasons: (1) the GoPro camera lacks a depth 
sensor, and (2) IR-based depth sensor rays may scatter off regolith particles, 
introducing noise that may corrupt splat training. 

\subsection{Gaussian Splatting}

\paragraph{Variants}
This paper experiments with six 3DGS variants, including vanilla
3DGS~\cite{kerbl20233d} as the baseline. 
3DGS represents a scene as an unstructured set of 3D Gaussian
primitives. Each
Gaussian contributes color and opacity to the final image through
a training process by
minimizing a base photometric loss by default:
\begin{equation}
\mathcal{L}_{base} = \lambda_{photo}\|C_{out} - C_{obs}\|_1 +
\lambda_{D\text{-}SSIM}\,\mathcal{L}_{D\text{-}SSIM}
\label{eq:perceptual_loss}
\end{equation}
where $C_{out}$ is the rendered RGB color at a pixel,
$C_{obs}$ is the corresponding observed pixel color,
$\|\cdot\|_1$ is the L1 photometric reconstruction error,
and $\mathcal{L}_{D\text{-}SSIM}$ penalizes perceptual differences
in local image structure. The scalars $\lambda_{photo}$ and
$\lambda_{D\text{-}SSIM}$ balance the two terms respectively.

While effective under ideal conditions, vanilla 3DGS assumes pinhole
cameras, clean imagery, dense Structure-from-Motion (SfM)
initialization, and uniform viewpoint sampling, assumptions that are
routinely violated in Lunar simulant testbeds. Also, regolith surfaces are largely featureless, causing standard
SfM to produce sparse or unreliable point clouds. \textbf{GaussianPro}~\cite{gspro} recovers
from this via Multi-View Stereo (MVS) and depth map-guided Gaussian propagation that
explicitly densifies low-texture regions. \textbf{Mip-Splatting}~\cite{yu2024mip}
eliminates aliasing artifacts that arise when the same regolith patch
is imaged from varying standoff distances during rover traverse.
\textbf{BAGS}~\cite{peng2024bags} handles motion blur introduced by
platform vibration using a blur proposal network.
\textbf{3DGRUT}~\cite{loccoz20243dgrt,wu20253dgut} corrects for
non-linear camera distortion in the wide-angle RealSense D456 mast camera.
Finally, \textbf{SparseGS}~\cite{sparsegs} regularizes reconstruction
from the limited viewpoint diversity of a rover traverse via
depth map (using DA3) priors and diffusion-based guidance, suppressing floater
artifacts in under-constrained regions.
SparseGS~\cite{sparsegs} is excluded from $\mathcal{R}_{\mathrm{Ctrl},i}$ as it is designed
for sparse-view conditions more representative of the rover modality. Overall, each of the six Gaussian Splat variants
evaluated targets a failure mode directly relevant
to Lunar terrain traversal.

\paragraph{Evaluation Metrics}
This work evaluates the NVS quality of the reconstructed splats
using the following three metrics on both scene-specific test-sets and cross-reconstruction PoVs,
following~\cite{zhang20253d}. Given a reference image $I$ and a 
rendered image $\hat{I}$ of height $H$ and width $W$, with maximum 
pixel intensity $L = 255$, \textbf{PSNR} quantifies pixel-level fidelity as the ratio of 
maximum signal power to mean reconstruction error:
\begin{equation}
\text{PSNR}(I, \hat{I}) = 10 \cdot \log_{10} \left( \frac{L^2}{\text{MSE}} \right)
\end{equation}
where $\text{MSE} = \frac{1}{HW}\sum_{i,j}\|I_{ij} - 
\hat{I}_{ij}\|^2$ is the mean squared error averaged over all 
pixels. Higher values indicate lower pixel-wise distortion.

\textbf{SSIM} captures perceptual quality by jointly assessing 
local luminance, contrast, and structure between image patches:
\begin{equation}
\text{SSIM}(I, \hat{I}) = \frac{(2\mu_I \mu_{\hat{I}} + 
C_1)(2\sigma_{I\hat{I}} + C_2)}
{(\mu_I^2 + \mu_{\hat{I}}^2 + C_1)(\sigma_I^2 + 
\sigma_{\hat{I}}^2 + C_2)}
\end{equation}
where $\mu_I$ and $\mu_{\hat{I}}$ are the local patch means of 
$I$ and $\hat{I}$, $\sigma_I^2$ and $\sigma_{\hat{I}}^2$ are 
their local variances, $\sigma_{I\hat{I}}$ is their local 
cross-covariance, and $C_1$, $C_2$ are small stability constants 
that prevent division by zero. Local patches are computed following\cite{kerbl20233d}, with a sliding $11\times11$ Gaussian window with a standard deviation of 1.5. SSIM $\in [-1,1]$, with values 
closer to 1 indicating greater structural agreement.

\textbf{LPIPS} measures perceptual distance using feature 
activations extracted by a pretrained deep neural network. For each 
layer $l$ of $L$ total layers, activations $\phi^l(I)$ and 
$\phi^l(\hat{I})$ are computed at each spatial location 
$(h, w)$ of the feature map of height $H_l$ and width $W_l$, 
weighted by learned channel coefficients $w_l$:
\begin{equation}
\text{LPIPS}(I, \hat{I}) = \sum_{l} \frac{1}{H_l W_l} 
\sum_{h,w} \left\| w_l \odot 
\left( \phi^l(I)_{hw} - \phi^l(\hat{I})_{hw} \right) \right\|_2^2
\end{equation}
where $\odot$ denotes element-wise multiplication. Lower LPIPS 
indicates higher perceptual similarity. 

\begin{algorithm}
\caption{Coverage Score ($S_c$)}
\label{alg:cs}
\begin{algorithmic}[1]
\Require COLMAP\cite{colmap} poses $\{\mathbf{R}_i,\mathbf{t}_i\}$;
         point cloud $\mathcal{P}$ with observer lists $\mathcal{O}(\mathbf{p})$
\Ensure $S_c\in[0,1]$
\State $d \gets \|\mathbf{C}_{(95)}-\mathbf{C}_{(5)}\|_2$
\State $W_f \gets \frac{1}{|\mathcal{P}|}\sum_{\mathbf{p}\in\mathcal{P}}
         \mathbf{1}\!\left[
           \max_{i,j\in\mathcal{O}(\mathbf{p})}
           \arccos\!\left(\hat{\mathbf{r}}_i^\top\hat{\mathbf{r}}_j\right)
           > \theta^{*}
         \right]$
\State $W_d \gets \bigl|\bigl\{\mathbf{p}\in\mathcal{P}:\theta(\mathbf{p})>\theta^{*}\bigr\}\bigr|/\ d^3$
\State $S_c \gets
         \left(\dfrac{W_f}{W_{f,\text{ref}}} \cdot \dfrac{W_d}{W_{d,\text{ref}}}\right)^{1/2}$
\State \Return $S_c$
\end{algorithmic}
\end{algorithm}
\vspace{-1.5em}
\subsection{Coverage Score ($S_{c}$)}
\label{sec:cs}

This paper proposes the \textbf{Coverage Score} $S_c\in[0,1]$ via two COLMAP\cite{colmap}-derived metrics: Wide-Angle Reconstruction Fraction ($W_f$) and Wide-Angle Point Density ($W_d$) (Algorithm~\ref{alg:cs}).

\paragraph{Wide-Angle Reconstruction Fraction ($W_f$)}
Let $\mathbf{R}_i\in SO(3)$ and $\mathbf{t}_i\in\mathbb{R}^3$ be the rotation
and translation of the $i$-th camera pose, $\mathbf{C}_i=-\mathbf{R}_i^\top\mathbf{t}_i$
its center, and
$d=\|\mathbf{C}_{(95)}-\mathbf{C}_{(5)}\|_2$
the cluster diagonal, where $\mathbf{C}_{(k)}$ denotes the per-axis $k$-th percentile
of $\{\mathbf{C}_i\}$ to ignore outliers.
Let $\mathcal{P}$ be the COLMAP\cite{colmap} point cloud and $\mathcal{O}(\mathbf{p})$ the
set of cameras observing point $\mathbf{p}$. The two farthest cameras that see $\mathbf{p}$ form the maximum pair-wise triangulation angle for the point.
$W_f$ is the fraction of points in $\mathcal{P}$ where this angle exceeds the threshold $\theta^{*}=30^{\circ}$. This threshold is chosen because a sweep of $\theta^{*}$
from $5^{\circ}$ to $60^{\circ}$ confirms that clean separation between
handheld (GoPro) and rover mast camera captures first emerges at
$25^{\circ}$ and is maximized at $30^{\circ}$. The angle is computed from unit rays
$\hat{\mathbf{r}}_i=(\mathbf{C}_i-\mathbf{p})/\|\mathbf{C}_i-\mathbf{p}\|$
between all observer pairs. Low $W_f$ indicates less novel views captured in the dataset. 
\paragraph{Wide-Angle Point Density ($W_d$)}
A single perfectly-observed point gives
$W_f=1$ regardless of how degenerate the reconstruction is.
$W_d$ remedies this by counting points with maximum triangulation angle $\theta(\mathbf{p})>\theta^{*}$
and normalizing by $d^3$.
Removing cameras that contribute
only low-angle observations reduces $W_d$ even when $W_f$ stays constant.

\paragraph{Coverage Score ($S_c$)}
Each metric is normalized to a reference calibrated on the best achieved values
($W_{f,\text{ref}}{=}0.143$ from GoPro Scene\,1;
 $W_{d,\text{ref}}{=}16.51$ from GoPro Scene\,2). Finally, $S_c$ is obtained by taking the geometric mean of these normalized values (Line\,5, Algorithm~\ref{alg:cs}).
$S_c{<}1$
reflects a proportional deficit.
For scenes notably larger than a $6\,\mathrm{m}\times4\,\mathrm{m}$ reference zone, the
reference values can be held fixed by partitioning the same into sub-zones of equivalent area.

\begin{table}
\centering
\caption{Rover-derived splat training benchmark across three Lunar analog scenes.
All models trained for 30,000 iterations at $819\times463$ pixels.
Values reported as mean\,$\pm$\,std.}
\label{tab:splats_mastcam}
\setlength{\tabcolsep}{3.5pt}
\begin{tabular}{l ccc}
\toprule
\textbf{Method} & $\mathrm{PSNR}_{Tr}\uparrow$ & $\mathrm{SSIM}_{Tr}\uparrow$ & $\mathrm{LPIPS}_{Tr}\downarrow$ \\
\midrule
\multicolumn{4}{l}{\textit{Scene\,1 \hfill 500 imgs (Train: 437; Test: 63)}} \\
\midrule
3DGS
  & 31.04{\scriptsize$\pm$2.51} & 0.88{\scriptsize$\pm$.04} & 0.30{\scriptsize$\pm$.05} \\
BAGS
  & 25.05{\scriptsize$\pm$2.41} & 0.72{\scriptsize$\pm$.06} & 0.29{\scriptsize$\pm$.04} \\
GaussianPro (DA3)
  & 30.15{\scriptsize$\pm$2.14} & 0.85{\scriptsize$\pm$.03} & 0.37{\scriptsize$\pm$.03} \\
GaussianPro (rend.)
  & 30.17{\scriptsize$\pm$2.14} & 0.85{\scriptsize$\pm$.03} & 0.37{\scriptsize$\pm$.03} \\
Mip-Splatting
  & 31.28{\scriptsize$\pm$2.49} & 0.89{\scriptsize$\pm$.04} & \textbf{0.25{\scriptsize$\pm$.05}} \\
3DGRUT
  & \textbf{33.26{\scriptsize$\pm$2.86}} & \textbf{0.92{\scriptsize$\pm$.04}} & 0.28{\scriptsize$\pm$.06} \\
SparseGS (full)
  & 30.29{\scriptsize$\pm$2.31} & 0.86{\scriptsize$\pm$.04} & 0.34{\scriptsize$\pm$.04} \\
SparseGS (20\%)
  & 30.08{\scriptsize$\pm$1.77} & 0.85{\scriptsize$\pm$.03} & 0.33{\scriptsize$\pm$.03} \\
\cmidrule(r){1-4}
\multicolumn{4}{r}{\scriptsize\textit{Scene Coverage:} $W_f{=}3.5\%,\ W_d{=}0.57,\ S_c{=}0.092$} \\
\midrule
\multicolumn{4}{l}{\textit{Scene\,2 \hfill 475 imgs (Train: 415; Test: 60)}} \\
\midrule
3DGS
  & 28.79{\scriptsize$\pm$2.13} & 0.87{\scriptsize$\pm$.03} & \textbf{0.23{\scriptsize$\pm$.03}} \\
BAGS
  & 20.54{\scriptsize$\pm$1.16} & 0.52{\scriptsize$\pm$.04} & 0.36{\scriptsize$\pm$.03} \\
GaussianPro (DA3)
  & 27.16{\scriptsize$\pm$1.75} & 0.82{\scriptsize$\pm$.04} & 0.32{\scriptsize$\pm$.03} \\
GaussianPro (rend.)
  & 27.21{\scriptsize$\pm$1.75} & 0.82{\scriptsize$\pm$.04} & 0.32{\scriptsize$\pm$.03} \\
Mip-Splatting
  & 28.28{\scriptsize$\pm$2.46} & 0.87{\scriptsize$\pm$.04} & 0.23{\scriptsize$\pm$.05} \\
3DGRUT
  & \textbf{30.19{\scriptsize$\pm$1.97}} & \textbf{0.90{\scriptsize$\pm$.02}} & 0.26{\scriptsize$\pm$.03} \\
SparseGS (full)
  & 27.84{\scriptsize$\pm$1.99} & 0.85{\scriptsize$\pm$.04} & 0.27{\scriptsize$\pm$.04} \\
SparseGS (20\%)
  & 26.38{\scriptsize$\pm$2.50} & 0.80{\scriptsize$\pm$.06} & 0.30{\scriptsize$\pm$.04} \\
\cmidrule(r){1-4}
\multicolumn{4}{r}{\scriptsize\textit{Scene Coverage:} $W_f{=}1.7\%,\ W_d{=}0.18,\ S_c{=}0.036$} \\
\midrule
\multicolumn{4}{l}{\textit{Scene\,3 \hfill 489 imgs (Train: 427; Test: 62)}} \\
\midrule
3DGS
  & 28.45{\scriptsize$\pm$2.37} & 0.88{\scriptsize$\pm$.04} & \textbf{0.26{\scriptsize$\pm$.03}} \\
BAGS
  & 22.76{\scriptsize$\pm$1.57} & 0.64{\scriptsize$\pm$.07} & 0.34{\scriptsize$\pm$.02} \\
GaussianPro (DA3)
  & 27.53{\scriptsize$\pm$1.87} & 0.84{\scriptsize$\pm$.04} & 0.36{\scriptsize$\pm$.03} \\
GaussianPro (rend.)
  & 27.56{\scriptsize$\pm$1.77} & 0.85{\scriptsize$\pm$.04} & 0.35{\scriptsize$\pm$.03} \\
Mip-Splatting
  & 27.46{\scriptsize$\pm$2.79} & 0.87{\scriptsize$\pm$.04} & 0.27{\scriptsize$\pm$.04} \\
3DGRUT
  & \textbf{29.73{\scriptsize$\pm$2.50}} & \textbf{0.92{\scriptsize$\pm$.02}} & 0.28{\scriptsize$\pm$.03} \\
SparseGS (full)
  & 28.18{\scriptsize$\pm$1.92} & 0.87{\scriptsize$\pm$.04} & 0.31{\scriptsize$\pm$.03} \\
SparseGS (20\%)
  & 25.71{\scriptsize$\pm$3.20} & 0.82{\scriptsize$\pm$.07} & 0.32{\scriptsize$\pm$.05} \\
\cmidrule(r){1-4}
\multicolumn{4}{r}{\scriptsize\textit{Scene Coverage:} $W_f{=}4.7\%,\ W_d{=}0.83,\ S_c{=}0.129$} \\
\bottomrule
\end{tabular}
\setlength{\tabcolsep}{6pt}
\vspace{-1.5em}
\end{table}

\subsection{Cross-Reconstruction Photorealistic NVS Fidelity (CRPNVS)}

CRPNVS fidelity is defined as the photometric accuracy measured by SSIM, PSNR,
and LPIPS of a \textit{reconstructed} scene, evaluated by comparing its rendered
viewpoints against matching and novel physical ground truth and/or a human-controlled
baseline reconstruction, where the baseline is free of traversal effects. For these matching PoVs to be extracted from the reconstruction being tested, it needs to be superimposed on the human-controlled reconstruction manually or using reference markers. 

For $\mathcal{R}_{\mathrm{Ctrl},i}$, the held-out test metrics are directly
interpretable as CRPNVS metrics, as the ``N-shaped'' sweep produces minimal trajectory
bias (verified by high $W_f$, Table~\ref{tab:splats_gopro}), and the
test-set contains undisturbed physical viewpoints. Neither condition
holds for $\mathcal{R}_{\mathrm{Rov},i}$, which additionally exhibits surface
warping. For example, in Fig. \ref{fig:crpnvs}, it can be seen that some 3D Gaussians form lofted from the surface, creating visual floaters for $\mathcal{R}_{\mathrm{Rov}}$. In such cases, a single alignment from fiducial markers might be significantly noisy and thus requires an averaged analysis across multiple small offsets about the datum level (the exact ground plane matched from the fiducial markers). This research deals with these issues in four-stages:

\begin{figure}[!htp]
\vspace{-0.5em}
    \centering
    \includegraphics[width=1\linewidth, angle=0]{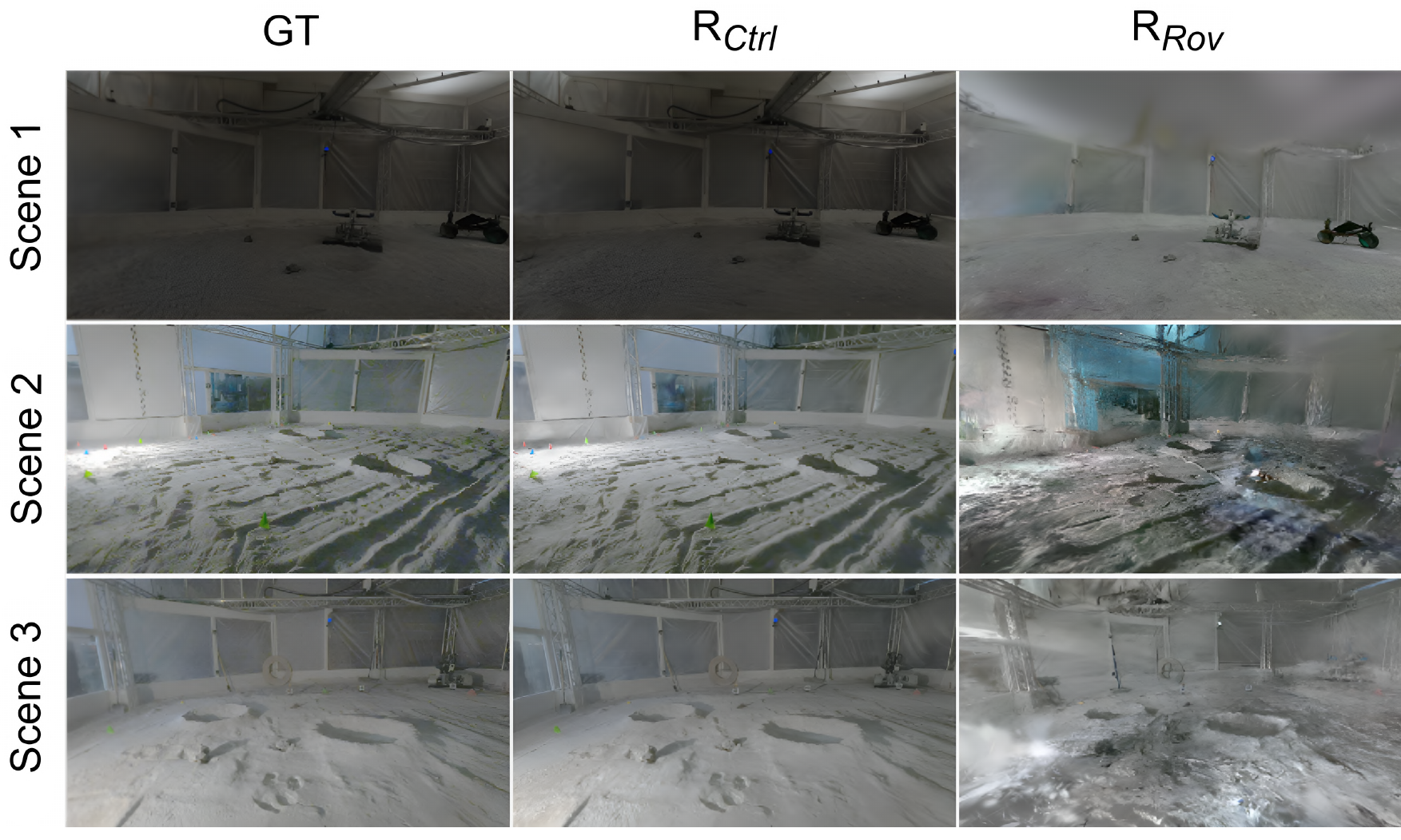}
    \caption{A single alignment from the three-panel comparison used in CRPNVS assessment. $\mathcal{R}_{\mathrm{Rov}}$ splat shows visible floaters on top of the matched ground plane.}
    \label{fig:crpnvs}
\vspace{-1.0em}
\end{figure}

\begin{enumerate}
    \item \textbf{Alignment:} The best-performing $\mathcal{R}_{\mathrm{Ctrl},i}$ and all
    $\mathcal{R}_{\mathrm{Rov},i}$ splat variants are brought into a common metric coordinate frame
    via fiducial marker (ArUco, AprilTags) correspondences
    \cite{garrido2014automatic}, using a RANSAC-based Umeyama similarity
    transform \cite{umeyama1991least}. It recovers scale, rotation, and
    translation between the two independent COLMAP \cite{colmap}
    reconstructions, with physical scale recovered from known marker
    dimensions. Scale offset is compensated by true scale derived from the markers.

    \item \textbf{Warp compensation:} The $\mathcal{R}_{\mathrm{Rov},i}$ splat variants from the previous stage are aligned geometrically in  COLMAP~\cite{colmap} frame, but the floating 3D Gaussians cause visual noise and artifacts that visually spawn at some offset from the matched datum level.  To compensate, seven elevation offsets within $[-0.3,+0.3] \,\text{m}$ at
    $0.1\,\text{m}$ intervals are taken about and including the datum level and each paired with the $\mathcal{R}_{\mathrm{Ctrl},i}$. Hence, the total number of images to compare increases to $3\times7\times$ number of render points instead of only the number of held out test images (See Table\ref{tab:mastcam_crpnvs}).

    \item \textbf{Matched rendering:} Once aligned, for the same camera position, similar PoVs are rendered in both $\mathcal{R}_{\mathrm{Ctrl},i}$ and $\mathcal{R}_{\mathrm{Rov},i}$ splat variants. All camera positions are extracted from the human-controlled
    COLMAP~\cite{colmap} database, yielding three panels per viewpoint:
    physical ground truth, $\mathcal{R}_{\mathrm{Ctrl},i}$ render, and $\mathcal{R}_{\mathrm{Rov},i}$ render.

    \item \textbf{Metric computation:} SSIM, PSNR, and LPIPS are
    re-computed with mean and variance analysis across the three-panel comparison,
    producing the final CRPNVS metrics for the $\mathcal{R}_{\mathrm{Rov},i}$.
\end{enumerate}
\subsection{USDZ Asset Generation for Digital Twins}
The best-performing splat (ideally $\mathcal{R}_{\mathrm{Ctrl},i}$) based on the CRPNVS benchmarks is converted to USDZ via NVIDIA NuRec~\cite{nvidia_nurec_how_it_works} and 3DGRUT~\cite{loccoz20243dgrt,wu20253dgut} and loaded into Isaac Sim 5. This is complemented with a surface aligned SuGAR-derived mesh \cite{guedon2024sugar} as a collider, making the photorealistic reconstruction simulation-ready as a Lunar-analog DT asset.

\begin{table*}
\centering
\caption{$\mathcal{R}_{\mathrm{Rov}}$ CRPNVS benchmark with frame resolution $960\times540$ pixels. Values reported as mean\,$\pm$\,std.}
\label{tab:mastcam_crpnvs}
\setlength{\tabcolsep}{3.5pt}
\begin{tabular}{l @{\hspace{15pt}} cc @{\hspace{15pt}} cc @{\hspace{15pt}} cc}
\toprule
& \multicolumn{2}{c}{$\mathrm{PSNR}\uparrow$}
& \multicolumn{2}{c}{$\mathrm{SSIM}\uparrow$}
& \multicolumn{2}{c}{$\mathrm{LPIPS}\downarrow$} \\
\cmidrule(lr){2-3}\cmidrule(lr){4-5}\cmidrule(lr){6-7}
\textbf{Method}
  & vs\,$\mathcal{R}_{\mathrm{Ctrl}}$ & vs\,$\mathrm{GT}$
  & vs\,$\mathcal{R}_{\mathrm{Ctrl}}$ & vs\,$\mathrm{GT}$
  & vs\,$\mathcal{R}_{\mathrm{Ctrl}}$ & vs\,$\mathrm{GT}$ \\
\midrule
\multicolumn{7}{l}{\textit{Scene\,1 -- Diffused Indoor Light \hfill CRPNVS: $3\times 900$ imgs $\times\,7$ elevation offsets}} \\
\midrule
3DGS
  & 12.48\,{\scriptsize$\pm$1.45} & 12.31\,{\scriptsize$\pm$1.42}
  & 0.52\,{\scriptsize$\pm$.11} & 0.48\,{\scriptsize$\pm$.10}
  & 0.59\,{\scriptsize$\pm$.09} & 0.66\,{\scriptsize$\pm$.09} \\
BAGS
  & \textbf{13.29\,{\scriptsize$\pm$1.18}} & \textbf{13.11\,{\scriptsize$\pm$1.30}}
  & \textbf{0.59\,{\scriptsize$\pm$.06}} & \textbf{0.54\,{\scriptsize$\pm$.06}}
  & \textbf{0.56\,{\scriptsize$\pm$.10}} & \textbf{0.64\,{\scriptsize$\pm$.09}} \\
GaussianPro (DA3)
  & 12.06\,{\scriptsize$\pm$1.45} & 11.96\,{\scriptsize$\pm$1.42}
  & 0.49\,{\scriptsize$\pm$.09} & 0.45\,{\scriptsize$\pm$.12}
  & 0.62\,{\scriptsize$\pm$.06} & 0.71\,{\scriptsize$\pm$.08} \\
GaussianPro (rend.)
  & 12.11\,{\scriptsize$\pm$2.00} & 11.95\,{\scriptsize$\pm$1.57}
  & 0.49\,{\scriptsize$\pm$.09} & 0.45\,{\scriptsize$\pm$.11}
  & 0.62\,{\scriptsize$\pm$.06} & 0.71\,{\scriptsize$\pm$.08} \\
Mip-Splatting
  & 12.23\,{\scriptsize$\pm$1.38} & 12.06\,{\scriptsize$\pm$1.37}
  & 0.50\,{\scriptsize$\pm$.11} & 0.46\,{\scriptsize$\pm$.09}
  & 0.61\,{\scriptsize$\pm$.09} & 0.69\,{\scriptsize$\pm$.09} \\
3DGRUT
  & 12.54\,{\scriptsize$\pm$1.31} & 12.36\,{\scriptsize$\pm$1.30}
  & 0.55\,{\scriptsize$\pm$.09} & 0.50\,{\scriptsize$\pm$.08}
  & 0.62\,{\scriptsize$\pm$.07} & 0.68\,{\scriptsize$\pm$.06} \\
SparseGS (full)
  & 12.15\,{\scriptsize$\pm$1.45} & 11.99\,{\scriptsize$\pm$1.45}
  & 0.50\,{\scriptsize$\pm$.13} & 0.46\,{\scriptsize$\pm$.11}
  & 0.60\,{\scriptsize$\pm$.10} & 0.68\,{\scriptsize$\pm$.09} \\
SparseGS (20\%)
  & 12.77\,{\scriptsize$\pm$2.12} & 12.60\,{\scriptsize$\pm$2.11}
  & 0.49\,{\scriptsize$\pm$.15} & 0.45\,{\scriptsize$\pm$.13}
  & 0.68\,{\scriptsize$\pm$.09} & 0.73\,{\scriptsize$\pm$.08} \\
\midrule
\multicolumn{7}{l}{\textit{Scene\,2 -- South Pole, Minimal Dust \hfill CRPNVS: $3\times 729$ imgs $\times\,7$ elevation offsets}} \\
\midrule
3DGS
  & 14.76\,{\scriptsize$\pm$1.79} & 14.55\,{\scriptsize$\pm$1.80}
  & 0.54\,{\scriptsize$\pm$.08} & 0.44\,{\scriptsize$\pm$.06}
  & 0.59\,{\scriptsize$\pm$.05} & 0.71\,{\scriptsize$\pm$.07} \\
BAGS
  & \textbf{15.72\,{\scriptsize$\pm$1.75}} & \textbf{15.49\,{\scriptsize$\pm$1.71}}
  & 0.54\,{\scriptsize$\pm$.09} & \textbf{0.44\,{\scriptsize$\pm$.06}}
  & \textbf{0.59\,{\scriptsize$\pm$.05}} & \textbf{0.71\,{\scriptsize$\pm$.06}} \\
GaussianPro (DA3)
  & 13.90\,{\scriptsize$\pm$1.89} & 13.73\,{\scriptsize$\pm$1.90}
  & \textbf{0.56\,{\scriptsize$\pm$.09}} & 0.45\,{\scriptsize$\pm$.07}
  & 0.66\,{\scriptsize$\pm$.06} & 0.79\,{\scriptsize$\pm$.06} \\
GaussianPro (rend.)
  & 14.07\,{\scriptsize$\pm$1.99} & 13.89\,{\scriptsize$\pm$1.98}
  & 0.56\,{\scriptsize$\pm$.09} & 0.45\,{\scriptsize$\pm$.07}
  & 0.66\,{\scriptsize$\pm$.06} & 0.79\,{\scriptsize$\pm$.06} \\
Mip-Splatting
  & 14.50\,{\scriptsize$\pm$1.98} & 14.32\,{\scriptsize$\pm$1.96}
  & 0.52\,{\scriptsize$\pm$.09} & 0.43\,{\scriptsize$\pm$.07}
  & 0.61\,{\scriptsize$\pm$.05} & 0.74\,{\scriptsize$\pm$.07} \\
3DGRUT
  & 14.57\,{\scriptsize$\pm$1.58} & 14.33\,{\scriptsize$\pm$1.57}
  & 0.55\,{\scriptsize$\pm$.10} & 0.45\,{\scriptsize$\pm$.08}
  & 0.63\,{\scriptsize$\pm$.04} & 0.72\,{\scriptsize$\pm$.05} \\
SparseGS (full)
  & 14.18\,{\scriptsize$\pm$1.93} & 13.98\,{\scriptsize$\pm$1.89}
  & 0.51\,{\scriptsize$\pm$.11} & 0.42\,{\scriptsize$\pm$.08}
  & 0.61\,{\scriptsize$\pm$.06} & 0.74\,{\scriptsize$\pm$.07} \\
SparseGS (20\%)
  & 13.49\,{\scriptsize$\pm$2.67} & 13.34\,{\scriptsize$\pm$2.59}
  & 0.47\,{\scriptsize$\pm$.10} & 0.39\,{\scriptsize$\pm$.08}
  & 0.64\,{\scriptsize$\pm$.06} & 0.74\,{\scriptsize$\pm$.06} \\
\midrule
\multicolumn{7}{l}{\textit{Scene\,3 -- South Pole, Extreme Varying Dust \hfill CRPNVS: $3\times 961$ imgs $\times\,7$ elevation offsets}} \\
\midrule
3DGS
  & 15.96\,{\scriptsize$\pm$3.27} & 15.80\,{\scriptsize$\pm$3.16}
  & \textbf{0.69\,{\scriptsize$\pm$.09}} & \textbf{0.57\,{\scriptsize$\pm$.08}}
  & 0.57\,{\scriptsize$\pm$.08} & 0.73\,{\scriptsize$\pm$.08} \\
BAGS
  & \textbf{16.32\,{\scriptsize$\pm$2.55}} & \textbf{16.18\,{\scriptsize$\pm$2.51}}
  & 0.64\,{\scriptsize$\pm$.09} & 0.53\,{\scriptsize$\pm$.08}
  & \textbf{0.53\,{\scriptsize$\pm$.06}} & \textbf{0.70\,{\scriptsize$\pm$.08}} \\
GaussianPro (DA3)
  & 14.96\,{\scriptsize$\pm$2.92} & 14.83\,{\scriptsize$\pm$2.84}
  & 0.68\,{\scriptsize$\pm$.08} & 0.56\,{\scriptsize$\pm$.07}
  & 0.64\,{\scriptsize$\pm$.07} & 0.80\,{\scriptsize$\pm$.06} \\
GaussianPro (rend.)
  & 14.88\,{\scriptsize$\pm$3.02} & 14.74\,{\scriptsize$\pm$2.93}
  & 0.67\,{\scriptsize$\pm$.09} & 0.55\,{\scriptsize$\pm$.08}
  & 0.64\,{\scriptsize$\pm$.07} & 0.79\,{\scriptsize$\pm$.06} \\
Mip-Splatting
  & 16.00\,{\scriptsize$\pm$2.93} & 15.87\,{\scriptsize$\pm$2.85}
  & 0.66\,{\scriptsize$\pm$.09} & 0.54\,{\scriptsize$\pm$.07}
  & 0.57\,{\scriptsize$\pm$.07} & 0.75\,{\scriptsize$\pm$.08} \\
3DGRUT
  & 15.53\,{\scriptsize$\pm$2.89} & 15.40\,{\scriptsize$\pm$2.81}
  & \textbf{0.69\,{\scriptsize$\pm$.08}} & \textbf{0.57\,{\scriptsize$\pm$.08}}
  & 0.59\,{\scriptsize$\pm$.07} & 0.72\,{\scriptsize$\pm$.06} \\
SparseGS (full)
  & 16.14\,{\scriptsize$\pm$3.01} & 15.97\,{\scriptsize$\pm$2.95}
  & 0.68\,{\scriptsize$\pm$.09} & 0.56\,{\scriptsize$\pm$.07}
  & 0.58\,{\scriptsize$\pm$.08} & 0.75\,{\scriptsize$\pm$.07} \\
SparseGS (20\%)
  & 15.65\,{\scriptsize$\pm$3.15} & 15.51\,{\scriptsize$\pm$3.13}
  & 0.65\,{\scriptsize$\pm$.11} & 0.54\,{\scriptsize$\pm$.09}
  & 0.60\,{\scriptsize$\pm$.07} & 0.75\,{\scriptsize$\pm$.07} \\
\bottomrule
\vspace{-1em}
\end{tabular}
\end{table*}

\begin{table}
\vspace{-1.5em}
\centering
\caption{Splat Training and CRPNVS benchmark across three Lunar analog scenes (Human-Control) compared against physical ground truth.
All models trained for 30,000 iterations at ${\sim}960\times540$ pixels.
Values reported as mean\,$\pm$\,std.}
\label{tab:splats_gopro}
\setlength{\tabcolsep}{3.5pt}
\begin{tabular}{l ccc}
\toprule
\textbf{Method} & $\mathrm{PSNR}_{Tr}\uparrow$ & $\mathrm{SSIM}_{Tr}\uparrow$ & $\mathrm{LPIPS}_{Tr}\downarrow$ \\
\midrule
\multicolumn{4}{l}{\textit{Scene\,1 \hfill 900 imgs (Train: 787; Test: 113)}} \\
\midrule
3DGS~\cite{kerbl20233d}
  & 32.48{\scriptsize$\pm$3.57} & 0.92{\scriptsize$\pm$0.06} & 0.22{\scriptsize$\pm$0.09} \\
BAGS~\cite{peng2024bags}
  & 31.35{\scriptsize$\pm$3.36} & 0.91{\scriptsize$\pm$0.06} & 0.23{\scriptsize$\pm$0.10} \\
GaussianPro (DA3)~\cite{gspro}
  & 31.67{\scriptsize$\pm$3.30} & 0.91{\scriptsize$\pm$0.06} & 0.23{\scriptsize$\pm$0.09} \\
GaussianPro (rend.)~\cite{gspro}
  & 31.66{\scriptsize$\pm$3.26} & 0.91{\scriptsize$\pm$0.06} & 0.23{\scriptsize$\pm$0.09} \\
Mip-Splatting~\cite{yu2024mip}
  & \textbf{32.56}{\scriptsize$\pm$3.56} & \textbf{0.92}{\scriptsize$\pm$0.06} & \textbf{0.20}{\scriptsize$\pm$0.09} \\
3DGRUT~\cite{loccoz20243dgrt,wu20253dgut}
  & 31.77{\scriptsize$\pm$3.36} & 0.91{\scriptsize$\pm$0.06} & 0.29{\scriptsize$\pm$0.10} \\
\cmidrule(r){1-4}
\multicolumn{4}{r}{\scriptsize\textit{Scene Coverage:} $W_f{=}14.3\%,\ W_d{=}15.60,\ S_c{=}0.97$} \\
\midrule
\multicolumn{4}{l}{\textit{Scene\,2 \hfill 729 imgs (Train: 637; Test: 92)}} \\
\midrule
3DGS~\cite{kerbl20233d}
  & 32.13{\scriptsize$\pm$2.13} & 0.89{\scriptsize$\pm$0.08} & 0.32{\scriptsize$\pm$0.09} \\
BAGS~\cite{peng2024bags}
  & 31.32{\scriptsize$\pm$1.98} & 0.87{\scriptsize$\pm$0.08} & 0.35{\scriptsize$\pm$0.09} \\
GaussianPro (DA3)~\cite{gspro}
  & 32.00{\scriptsize$\pm$2.13} & 0.89{\scriptsize$\pm$0.08} & 0.33{\scriptsize$\pm$0.10} \\
GaussianPro (rend.)~\cite{gspro}
  & 32.02{\scriptsize$\pm$2.07} & 0.89{\scriptsize$\pm$0.08} & 0.33{\scriptsize$\pm$0.10} \\
Mip-Splatting~\cite{yu2024mip}
  & 32.35{\scriptsize$\pm$2.75} & \textbf{0.89}{\scriptsize$\pm$0.08} & \textbf{0.31}{\scriptsize$\pm$0.09} \\
3DGRUT~\cite{loccoz20243dgrt,wu20253dgut}
  & \textbf{32.53}{\scriptsize$\pm$2.09} & 0.88{\scriptsize$\pm$0.07} & 0.42{\scriptsize$\pm$0.09} \\
\cmidrule(r){1-4}
\multicolumn{4}{r}{\scriptsize\textit{Scene Coverage:} $W_f{=}11.7\%,\ W_d{=}16.51,\ S_c{=}0.91$} \\
\midrule
\multicolumn{4}{l}{\textit{Scene\,3 \hfill 961 imgs (Train: 840; Test: 121)}} \\
\midrule
3DGS~\cite{kerbl20233d}
  & \textbf{29.93}{\scriptsize$\pm$3.02} & 0.81{\scriptsize$\pm$0.08} & 0.39{\scriptsize$\pm$0.09} \\
BAGS~\cite{peng2024bags}
  & 28.77{\scriptsize$\pm$2.70} & 0.78{\scriptsize$\pm$0.09} & 0.43{\scriptsize$\pm$0.09} \\
GaussianPro (DA3)~\cite{gspro}
  & 29.47{\scriptsize$\pm$2.93} & 0.81{\scriptsize$\pm$0.08} & 0.41{\scriptsize$\pm$0.09} \\
GaussianPro (rend.)~\cite{gspro}
  & 29.38{\scriptsize$\pm$2.96} & 0.80{\scriptsize$\pm$0.08} & 0.41{\scriptsize$\pm$0.09} \\
Mip-Splatting~\cite{yu2024mip}
  & 29.80{\scriptsize$\pm$3.72} & \textbf{0.82}{\scriptsize$\pm$0.08} & \textbf{0.37}{\scriptsize$\pm$0.09} \\
3DGRUT~\cite{loccoz20243dgrt,wu20253dgut}
  & 29.91{\scriptsize$\pm$2.83} & 0.81{\scriptsize$\pm$0.08} & 0.47{\scriptsize$\pm$0.09} \\
\cmidrule(r){1-4}
\multicolumn{4}{r}{\scriptsize\textit{Scene Coverage:} $W_f{=}6.3\%,\ W_d{=}11.56,\ S_c{=}0.56$} \\
\bottomrule
\end{tabular}
\setlength{\tabcolsep}{6pt}
\vspace{-1.5em}
\end{table}

\begin{table}
\centering
\caption{COLMAP\cite{colmap} scale accuracy from triangulated fiducial markers.}
\label{tab:scale_accuracy}
\renewcommand{\arraystretch}{0.9}
\setlength{\tabcolsep}{4pt}
\begin{tabular}{cccccc}
\toprule
Scene & Modality & Tag & $\#$ Markers & Error & Spread \\
\midrule
\multirow{2}{*}{1} & $\mathcal{R}_{\mathrm{Ctrl},i}$ & \multirow{2}{*}{ArUco 4×4} & 5/6 & $+0.58\%$ & $2.85\%$ \\
                   & $\mathcal{R}_{\mathrm{Rov}}$ &                             & 3/6 & $+0.42\%$ & $23.33\%$ \\
\midrule
\multirow{2}{*}{2} & $\mathcal{R}_{\mathrm{Ctrl},i}$ & \multirow{2}{*}{ArUco 4×4} & 5/6 & $-3.22\%$ & $16.61\%$ \\
                   & $\mathcal{R}_{\mathrm{Rov}}$ &                             & 3/6 & $+7.57\%$ & $22.15\%$ \\
\midrule
\multirow{2}{*}{3} & $\mathcal{R}_{\mathrm{Ctrl},i}$ & \multirow{2}{*}{AprilTag 36h11}  & 3/4 & $-0.04\%$ & $2.75\%$ \\
                   & $\mathcal{R}_{\mathrm{Rov}}$ &                             & 1/4 & $-0.18\%$ & $14.38\%$ \\
\bottomrule
\end{tabular}
\vspace{-1.5em}
\end{table}

\section{Results and Discussion}
This section highlights the results showing why CRPNVS metrics should be prioritized over training metrics, the fidelity gap between human and rover-based reconstructions and how $S_c$ can be used as a pre-diagnostic heuristic in the absence of baselines.
\paragraph{\textbf{Training vs CRPNVS Benchmark}}
BAGS\cite{peng2024bags} performs the overall worst in $\mathcal{R}_{\mathrm{Rov},i}$ training (Table \ref{tab:splats_mastcam}) but the overall best in CRPNVS metrics (Table \ref{tab:mastcam_crpnvs}). This verifies the central claim that training metrics do not represent cross-reconstruction photorealistic accuracy under traversal effects. A linear trajectory bias causes held-out
test views to be near-identical to training views, inflating training metrics without reflecting generalization to novel viewpoints. Additionally, view degrading factors (dusty PoVs and motion blur) may deviate the experienced environments far enough from GTs to invalidate training method ranking. Thus, for photorealistic scene reconstructions with data under traversal effects, CRPNVS metrics should be prioritized. From the CRPNVS results, \textit{this paper recommends BAGS\cite{peng2024bags} as the recommended method for scene reconstruction under traversal effects}, as its blur proposal network can better address the blur and haze in Lunar-analogs produced by rover motion and lofted regolith. Additionally, if the rover data are extremely sparse, DA3-guided SparseGS\cite{sparsegs} could be considered as it can mostly retain similar levels of CRPNVS performance even with 20\% of the full data. \textit{For human-controlled imagery, Mip-Splatting\cite{yu2024mip} is the recommended method}, as in undisturbed scenes (Table \ref{tab:splats_gopro}), its anti-aliasing filter produces more photorealistic results close to GT. 
\paragraph{\textbf{Human-Controlled vs Rover-Derived Reconstruction}} From the CRPNVS metrics between $\mathcal{R}_{\mathrm{Ctrl},i}$ and $\mathcal{R}_{\mathrm{Rov},i}$, rover-derived  $\mathcal{R}_{\mathrm{Rov},i}$ undergoes an average drop of 46--60\%  in PSNR, 31--51\% in SSIM, and an 89--220\% increase in LPIPS. Table \ref{tab:scale_accuracy} also shows significantly higher scale ambiguity in $\mathcal{R}_{\mathrm{Rov},i}$ compared to $\mathcal{R}_{\mathrm{Ctrl},i}$.
This quantifies the superiority of the proposed human-controlled reconstruction in terms of both photorealism and scale accuracy, and the expected tradeoff when considering rover-derived data that has undergone traversal effects.

\begin{figure}
    \centering
    \includegraphics[width=1\linewidth]{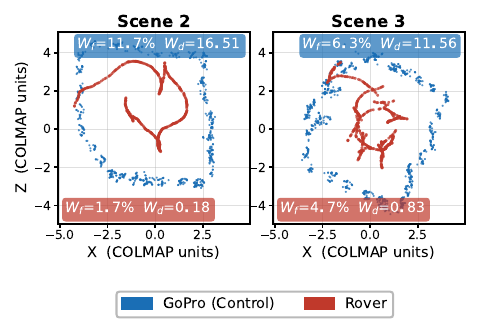}
    \caption{Scene 3's less linear rover camera traverse yields higher $W_{f}$ and $W_{d}$ than scene 2 despite harsher dust conditions, explaining its superior CRPNVS SSIM.}
    \label{fig:trajectories}
    \vspace{-1.5em}
\end{figure}

\paragraph{\textbf{Coverage Score and CRPNVS}}
The six datasets (3 scenes $\times$ 2 sensors) show that a high $S_c$ heuristically corresponds to a high overall CRPNVS, evidenced by the large
disparity between $\mathcal{R}_{\mathrm{Ctrl},i}$ and $\mathcal{R}_{\mathrm{Rov},i}$. In fact, $S_c$ shows the same rank as the highest achievable CRPNVS SSIM
for $\mathcal{R}_{\mathrm{Ctrl}}$ against physical GTs.
Notably, $S_c$ overturns the presumption that Scene~3,
with its extreme lofted dust, should yield inferior CRPNVS SSIM to
Scene~2. This is because SSIM's spatial structure term is directly
sensitive to triangulation geometry. As shown in Fig.~\ref{fig:trajectories},
Scene~3 has substantially higher $W_f$ and $W_d$ than Scene~2,
reflecting better scene coverage leading to better SSIM. 
This might make $S_c$ a practical acquisition diagnostic as given a candidate
rover traverse plan, $S_c$ could estimate expected reconstruction structural
fidelity before any splat is trained and validated in the absence of a controlled baseline or clean GTs. It may also propose trajectories for optimizing scene reconstruction.

$S_c$ does not, however, scale with PSNR or LPIPS, nor can it
predict SparseGS behavior. PSNR reduces quality to a global average of
squared pixel error, while LPIPS operates on deep
perceptual features. SparseGS compounds this further by using diffusion
priors to hallucinate unobserved geometry. Thus, these three processes are beyond
what COLMAP\cite{colmap}-derived geometry can anticipate. 

\subsection{Limitations and Future Work}
This study has five principal limitations. First, broader scene diversity may be needed to further validate $S_{c}$. Readers are encouraged to consider it as a pilot study. Second, $S_{c}$ relies on existing points in COLMAP\cite{colmap} database and cannot relate to non-geometric metrics and diffusion-based view filling. Third, using solely monocular capture makes the pipeline highly reproducible but also introduces scale ambiguity, especially in the absence of fiducial markers. Currently scale accuracy is regained using fiducial markers. Future work will incorporate the overhead LiDARs at the MLSS to further assess geometric fidelity alongside photometric fidelity. Fourth, the study only included 3DGS variants that work in post-processing. Future work will emphasize the viability of near-real-time variants. Finally, despite the best efforts, there might be slight changes between the human-controlled and rover scans. For example, some distant lights being turned off mid-test during the rover scans while they were turned on during the human-controlled scans and vice-versa as the testbed is located in a shared facility. This might have caused minimal downgrade in $\mathcal{R}_{\mathrm{Rov},i}$ CRPNVS performance as light is diffused through the testbed walls and glass windows. Future evaluations will minimize such occurrences. Nevertheless, the internal setup with rocks, craters and lights was kept unchanged.

\section{Conclusion}
This work outlined a reproducible coverage-aware photorealistic scene reconstruction pipeline for Lunar-analog DTs. It also quantified that standard 3DGS training metrics under traversal effects are structurally incapable of validating photorealistic scene reconstruction fidelity by replicating the pipeline with rover data under traversal effects.
The CRPNVS evaluation framework resolved this by grounding post-reconstruction renders against physical captures at matched and clean novel viewpoints, making the fidelity gap between human-controlled and rover modalities directly measurable in Lunar-analog reconstructions.
The most actionable finding was the BAGS inversion. It ranked last in training metrics under traversal effects, but excelled in overall CRPNVS, confirming that blur-aware reconstruction generalized better under traversal degradation than methods optimized for held-out training-view fidelity.
In addition, preliminary results showed that the piloted Coverage Score, $S_c$, quantified scene coverage as a dominant predictor in reconstruction quality. It may provide a complementary pre-reconstruction heuristic computed purely from COLMAP~\cite{colmap} geometry in the absence of human-controlled baselines for CRPNVS.
Therefore, such traits might make it a practical acquisition diagnostic given an un-referenced candidate rover traverse plan. Validating this path across broader terrain configurations with near-real-time methods and integrating geometric fidelity via overhead LiDARs remain the immediate priorities for follow-on work.
\section*{Acknowledgments}
Claude Sonnet 4.6 and Overleaf AI were used to improve clarity and grammatical flow. All content was subsequently reviewed and approved by the authors, who bear full responsibility for its accuracy and integrity. The authors are also grateful to Dr. Chris Dreyer, Hans Mertens and their teams for supporting infrastructure.


\bibliographystyle{IEEEtran}
\bibliography{ref}
\end{document}